Multi-view information fusion using multi-view variational autoencoder to predict proximal femoral fracture load

Chen Zhao[1,#], Joyce H. Keyak[2,#], Xuewei Cao[3], Qiuying Sha[3], Li Wu[4], Zhe Luo[4], Lan-Juan Zhao[4], Qing Tian[4], Michael Serou[5], Chuan Qiu[4], Kuan-Jui Su[4], Hui Shen[4], Hong-Wen Deng[4,*], Weihua Zhou[1,6,*]

1. Department of Applied Computing, Michigan Technological University, 1400 Townsend Dr, Houghton, MI 49931, USA

2. Department of Radiological Sciences, Department of Biomedical Engineering, Department of Mechanical and Aerospace Engineering, and Chao Family Comprehensive Cancer Center, University of California, Irvine, Irvine, CA 92697, USA

3. Department of Mathematical Sciences, Michigan Technological University, 1400 Townsend Dr, Houghton, MI 49931, USA

4. Division of Biomedical Informatics and Genomics, Tulane Center of Biomedical Informatics and Genomics, Deming
Department of Medicine, Tulane University, New Orleans, LA 70112, USA

5. Department of Radiology, Deming Department of Medicine, School of Medicine, Tulane University, New Orleans, LA 70112, USA

6. Center for Biocomputing and Digital Health, Institute of Computing and Cybersystems, and Health Research Institute, Michigan Technological University, Houghton, MI 49931, USA

# Chen Zhao[1,#] and Joyce H Keyak[2,#] contributed equally to this work.

* Corresponding authors and lead contacts:

Weihua Zhou, Ph.D.

Department of Applied Computing, Michigan Technological University,

1400 Townsend Dr, Houghton, MI, 49931, USA

Tel: 906-487-2666

E-Mail: whzhou@mtu.edu

Hong-Wen Deng, Ph.D.

Tulane Center for Biomedical Informatics and Genomics, Deming Department of Medicine, Tulane University,

1440 Canal Street, Suite 1619F, New Orleans, LA 70112, USA

Tel: 504-988-1310

Email: hdeng2@tulane.edu

## Abstract

*Background and aim*: Hip fracture occurs when an applied force exceeds the force that the proximal femur can support (the fracture load or "strength") and can have devastating consequences with poor functional outcomes. Proximal femoral strengths for specific loading conditions can be computed by subject-specific finite element analysis (FEA) using quantitative computerized tomography (QCT) images. However, the cost, radiation, and availability of QCT limit its clinical usability. Alternative low-cost and widely available measurements, such as dual energy X-ray absorptiometry (DXA) and genetic factors, would be preferable for bone strength assessment. The aim of this paper is to design a deep learning-based model to predict proximal femoral strength using multi-view information fusion.

*Method*: We developed new models using multi-view variational autoencoder (MVAE) for feature representation learning and a product of expert (PoE) model for multi-view information fusion. We applied the proposed models to an in-house Louisiana Osteoporosis Study (LOS) cohort with 931 male subjects, including 345 African Americans and 586 Caucasians. With an analytical solution of the product of Gaussian distribution, we adopted variational inference to train the designed MVAE-PoE model to perform common latent feature extraction. We performed genome-wide association studies (GWAS) to select 256 genetic variants with the lowest p-values for each proximal femoral strength and integrated whole genome sequence (WGS) features and DXA-derived imaging features to predict proximal femoral strength.

*Results*: The best prediction model for fall fracture load was acquired by integrating WGS features and DXA-derived imaging features. The designed models achieved the mean absolute percentage error of 18.04%, 6.84% and 7.95% for predicting proximal femoral fracture loads using linear models of fall loading, nonlinear models of fall loading, and nonlinear models of stance loading, respectively. Compared to existing multi-view information fusion methods, the proposed MVAE-PoE achieved the best performance.

*Conclusion*: The proposed models are capable of predicting proximal femoral strength using WGS features and DXA-derived imaging features. Though this tool is not a substitute for FEA using QCT images, it would make improved assessment of hip fracture risk more widely available while avoiding the increased radiation dosage and clinical costs from QCT.

## 1. Introduction

The increasing elderly population and the rise in fracture incidence have made osteoporosis a considerable public health issue in U.S. Osteoporosis causes bones to become weak and brittle, leading to osteoporotic fractures. Osteoporosis affects about 18% of women and 6% of men globally [1]. The economic burden of osteoporosis has been estimated at between $17 billion and $20.3 billion (2020 data) [2]. Fracture of the proximal femur is a common and disastrous health outcome that limits previously functional elderly patients from living independently. Each year over 300,000 older people in the U.S. are hospitalized for hip fracture [3]. The reported mortality rate is up to 20-24% in the first year after a hip fracture [4,5], and a greater risk of dying may persist for at least 5 years [6]. An inexpensive and accurate prognostic instrument for hip fracture risk assessment would enable individuals with a high risk for osteoporotic hip fracture to receive preventative treatment [7].

For the diagnosis of osteoporosis, areal bone mineral density (aBMD), assessed by dual energy X-ray absorptiometry (DXA), is the standard diagnostic clinical parameter [8][9]. Although DXA-derived aBMD correlates with bone weakness and fragility fracture [10], DXA is a 2D-projection technique that poorly accounts for 3D bone geometry and size [11], while bone geometry and bone size have strong genetic determination [12,13]. Further, efforts toward dissecting the genetic basis of osteoporosis using genome-wide association studies (GWASs) have been mainly focused on aBMD traits which have been widely studied, but GWAS results only explain part of the variance in hip fracture risk [14]. Thus, both DXA-derived features and genetic factors provide limited information about skeletal factors on fracture risk. It has been shown that genetic determinants of aBMD, bone geometry and bone sizes are genetically correlated, sharing some commons genes [15].

Principles of physics dictate that hip fracture occurs when an applied force exceeds the force that the proximal femur can support. This force, the proximal femoral strength or fracture load, can be computed using subject-specific finite element analysis (FEA), which incorporates the biomechanically important features of the hip, i.e. the 3D bone geometry and distribution of bone density from quantitative computerized tomography (QCT) images [7,16–19]. Furthermore, FEA-computed proximal femoral strength is associated with incident hip fracture in men and women, and in men even after accounting for aBMD [17].

Although QCT-based FEA has shown significant value in the assessment of proximal femoral strength, the cost, radiation, and availability of QCT-based FEA limit its clinical usability. DXA images are less expensive and use less radiation, but only describe aBMD, and proximal femoral shape and size in 2D. However, for predicting bone strength, replacing 3D QCT with less robust 2D DXA data can potentially be compensated for by incorporating bone-strength related genetic variants [20]. Accordingly, DXA-derived imaging features and genetic features could be incorporated into prediction models with high clinical applicability. Further, prediction models would help integrate the large number of high-dimensional inter-correlated complicated predictors from genetic data and image features to draw an overall conclusion regarding proximal femoral strength in individual patients.

In this paper, we propose a novel model, multi-view variational autoencoder with the product of expert (MVAE-PoE) for proximal femoral strength prediction. The proposed MVAE-PoE incorporates variational autoencoder (VAE) to learn feature representation and employs the product of expert (PoE) for multi-view information fusion. A linear regression estimator is used to predict proximal femoral strength based on the extracted latent features. Extensive analyses were performed, leveraging the combination of whole genome sequence (WGS) data and DXA-derived imaging features.

## 2. Background and Related Work

*Prediction of proximal femoral strength.* QCT provides more accurate quantification of BMD in the lumbar spine and hip than DXA because QCT provides volumetric BMD (vBMD) while DXA calculates aBMD [21]. In addition, QCT-based FEA describes the hip mechanical behavior and provides more information about bone quality and fracture risk than DXA [7,16,18,19,22,23]. However, all QCT-based methods require the administration of ionizing radiation, which limit the use of QCT in clinical practice. Yang et al. demonstrated that supplementing standard DXA-derived aBMD measurements with sophisticated femoral trabecular bone characterization from DXA significantly improved the performance of predicting hip fracture load [24]. The aBMD measured by DXA is currently a standard clinical surrogate marker of bone strength to diagnose osteoporosis; however, integrating the heterogeneous distribution of bone material properties is more powerful for predicting bone strength [25].

Genetic markers are important for identifying subjects at risk of hip fracture through effects on proximal femoral strength/structure. Using WGS data, GWAS and large-scale collaborative studies have identified hundreds of genetic markers, explaining substantial proportions of population variation in osteoporotic traits [26], such as BMD [27–29] and fracture risk factors [30]. Though aBMD is an important phenotype that is clinically relevant to osteoporotic hip fracture, it explains limited variance in hip strength [14,31–34]. Therefore, it is important to discover how genetics influence FEA-computed proximal femoral strength, thereby influencing hip fracture risk.

*Multi-view information fusion using machine learning.* Multi-view learning (MVL) is a strategy for fusing data from different modalities. MVL approaches can be divided into three categories: 1) co-training; 2) multi-kernel learning; and 3) subspace learning [35]. Co-training exchanges the separately trained models and provides the pseudo labels with discriminative information between views to accelerate model training [36], which focuses more on semi-supervised learning tasks. Multi-kernel learning aims at finding the mapping between different views with different kernels, and then combines the projected features for information fusion [37]. Subspace learning aims at finding the common space between different views where the shared latent space contains the information of all views [38]. For subspace learning, autoencoder is widely used for extracting the common representation of the multi-view data [39]. We will focus on this type of method in this study.

*Variational autoencoder (VAE).* VAE, proposed by Kingma et al. [40], is a latent variable generative model which learns the deep representation of the input data. The goal of VAE is to maximize the marginal likelihood of the data (a.k.a evidence), which can be decomposed into a sum over marginal log-likelihoods of individual features, as illustrated in Eq. 1.

$$\log p_\theta(x^{(i)}) = D_{KL}\left(q_\phi(z|x^{(i)}) \parallel p_\theta(z|x^{(i)})\right) + \mathcal{L}(\theta, \phi; x^{(i)}) \tag{1}$$

where $x^{(i)}$ is the feature vector for $i$-th subject in the dataset $\{x^{(i)}\}_{i=1}^{N}$, $N$ is the number of subjects, $z$ is a random variable in the latent space, $q_\phi$ is the posterior approximation of $z$ with the learnable parameters $\phi$, $p_\theta$ is the ground truth posterior distribution of $z$ with the intractable parameters $\theta$, and $D_{KL}(\cdot \parallel \cdot)$ represents the Kullback–Leibler (KL) divergence between the approximated posterior distribution and the ground truth posterior distribution. Because of the non-negativity of the KL divergence, the log-likelihood $\log p_\theta(x^{(i)}) \geq \mathcal{L}(\theta, \phi; x^{(i)})$. If the approximated posterior distribution $q_\phi(z|x^{(i)})$ is identical to the ground truth posterior distribution $p_\theta(z|x^{(i)})$, then the $\log p_\theta(x^{(i)}) = \mathcal{L}(\theta, \phi; x^{(i)})$. Therefore, $\mathcal{L}(\theta, \phi; x^{(i)})$ is called the evidence lower bound (ELOB), which is defined by Eq. 2.

$$\begin{aligned}\mathcal{L}(\theta, \phi; x^{(i)}) &= \log p_\theta(x^{(i)}) - D_{KL}\left(q_\phi(z|x^{(i)}) \parallel p_\theta(z|x^{(i)})\right) \\ &= \mathbb{E}_{q_\phi(z|x^{(i)})}[\log p_\theta(x^{(i)}|z)] - D_{KL}\left(q_\phi(z|x^{(i)}) \parallel p_\theta(z|x^{(i)})\right)\end{aligned} \tag{2}$$

Thus, minimizing the KL divergence is equivalent to maximizing the ELOB. To train the model explicitly and implement the loss function in a closed form, we parameterize the $q_\phi$ as a multivariate normal distribution (multivariate Gaussian distribution) with an approximately diagonal variance-covariance matrix. Then the analytical solution for the KL divergence is shown in Eq. 3.

$$D_{KL}\left(q_\phi(z|x^{(i)}) \parallel p_\theta(z|x^{(i)})\right) = \frac{1}{2}\sum_{d=1}^{D}\left(\left(\mu_d^{(i)}\right)^2 + \left(\sigma_d^{(i)}\right)^2 - \log\left(\left(\sigma_d^{(i)}\right)^2\right) - 1\right) \tag{3}$$

where $D$ is the number of the latent variables extracted by the VAE, and $\mu_d^{(i)}$ and $\left(\sigma_d^{(i)}\right)^2$ are the approximate mean and variance of the posterior distribution of $d$-th latent variable for $i$-th subject.

## 3. Materials and methodology

### 3.1. Enrolled subjects and data generation

In this study, we propose the MVAE-PoE to predict the proximal femoral strength calculated from QCT-based FEA by integrating WGS features and DXA-derived imaging features. The studied cohort was acquired from the Louisiana Osteoporosis Study (LOS) [41,42]. The LOS cohort is an ongoing research dataset (>17,000 subjects accumulated so far) with recruitment starting in 2011, aimed at investigating both environmental and genetic risk factors for osteoporosis and other musculoskeletal diseases [43,44]. All participants signed an informed-consent document before any data collection, and the study was approved by the Tulane University Institutional Review Board. A total of 931 male subjects, consisting of 345 African Americans and 586 Caucasians with QCT images, WGS and DXA-derived features, were used in our analyses. The DXA-derived features were measured for each participant using a Hologic QDR-4500 Discovery DXA scanner (Hologic Inc., Bedford, MA, USA) by trained and certified research staff. To ensure quality assurance, the machine was calibrated daily using a phantom scan. The accuracy of BMD measurement was assessed by the coefficient of variation for repeated measurements, which was approximately 1.9% for femoral neck BMD [43]. The basic demographic information for the enrolled subjects is shown in Table 1.

**Table 1**. Demographic information for the enrolled subjects. The ranges are illustrated in the parenthesis. SD, standard deviation.

| Race | Age (year, mean±SD) | Height (cm) | Weight (kg) |
| --- | --- | --- | --- |
| African-American | 38.60±7.74 (20 to 51) | 174.69±7.03 (154.00 to 190.80) | 82.67±17.68 (50.80 to 135.20) |
| Caucasian | 35.22±8.53 (20 to 51) | 175.37±6.80 (154.94 to 198.00) | 83.25±16.07 (50.80 to 135.40) |
| All | 36.47±8.40 (20 to 51) | 175.12±6.89 (154.00 to 198.00) | 83.04±16.68 (50.80 to 135.40) |

*QCT image acquisition and FEA for calculation of proximal femoral strength.* The QCT scans (GE Discovery CT750 HD system; 2.5 mm-thick slices; pixel size, 0.695–0.986 mm; 512 × 512 matrix) were acquired at Tulane University Department of Radiology. For each QCT slice, contours of the left proximal femur were labeled by well-trained operators, in consultation with our experienced researcher (J.H.K). We developed in-house software for automated annotation, containing our previously developed deep learning-based segmentation model [45] and a thresholding algorithm with edge tracing [46], in combination with manual visualization and correction. Our deep learning model achieved a Dice similarity coefficient of 0.9888, indicating that only minor manual modifications are required for annotating new QCT images. Using the annotated contours, we then used linear and nonlinear FE models to estimate the strengths of the proximal femoral under two loading conditions, single-limb stance and loading from a fall onto the posterolateral aspect of the greater trochanter [19].

The nonlinear FEA models simulated mechanical testing of the femur in which displacement is incrementally applied to the femoral head [7,16–19]. The computed reaction force on the femoral head initially increases, reaches a peak value (the load capacity or fracture load), and then decreases. To achieve this mechanical behavior, the FEA models employ heterogeneous isotropic elastic moduli, yield strengths, and nonlinear post-yield properties. These properties are computed from the calibrated QCT density ($\rho_{CHA}$, g/cm³) of each voxel in an element, which are then used to compute the ash density ($\rho_{ash}$, g/cm³) ($\rho_{ash}$=0.0633+0.887 $\rho_{CHA}$), and $\rho_{ash}$ is used to compute mechanical properties. Each linear hexahedral finite element measures 2.5 mm on a side and the mechanical properties of the element are computed by averaging the values of each property over all voxels in the element, while accounting for the volume fraction of each voxel within the element. Together, these mechanical properties describe an idealized density-dependent nonlinear stress-strain curve for each element [16–19]. Material yield is defined to occur when the von Mises stress exceeds the yield strength of the element. After yield, the plastic flow was modeled assuming a plastic strain-rate vector normal to the von Mises yield surface and isotropic hardening/softening. Displacement is applied incrementally to the femoral head, and the reaction force on the femoral head is computed at each increment as the distal end of the model is fully constrained. For the fall models, the surface of the greater trochanter opposite the loaded surface of the femoral head was constrained in the direction of the displacements while allowing motion transversely. The nonlinear FEA-computed proximal femoral fracture load was defined as the maximum FEA-computed force on the femoral head, i.e., the load capacity.

For phenotypes, we calculated three proximal femoral strengths under the two loading conditions: linear fall fracture load (LF), nonlinear fall fracture load (NLF), and nonlinear stance fracture load (NLS). The LF represents the load at the onset of fracture [7]. To determine the LF, the factor of safety (FOS) at the centroid of each finite element in the model is calculated as the ratio of the yield strength of the finite element to the von Mises stress at the centroid of the element. The LF is defined as the force applied to the femoral head when the FOS values of 15 contiguous non-surface elements are equal to or less than 1.0 [7,19]. The NLF and NLS were calculated using the above-described method and represent the load capacity of the proximal femur (the maximum force of the femoral head can support) [19]. The basic statistical information for the calculated proximal femoral strengths is shown in Table 2.

**Table 2.** Statistical information for proximal femoral strengths under three loading conditions. The mean and standard deviation are illustrated in the parenthesis. N: Newton; SD, standard deviation; LF, linear fall fracture load; NLF, nonlinear fall fracture load; NLS, nonlinear stance fracture load.

| Race | LF (N, mean±SD) | NLF (N, mean±SD) | NLS (N, mean±SD) |
| --- | --- | --- | --- |
| African American | 2,587.16±831.52 (875 to 6,138) | 4,353.82±570.49 (2,555 to 5,823) | 21,414.04±4,350.22 (11,084 to 40,904) |
| Caucasian | 2,162.70±782.18 (716 to 7,247) | 4,281.34±562.30 (2,683 to 6,337) | 19,275.38±3,954.22 (10,092 to 32,818) |
| All | 2,319.99±826.24 (716 to 7,247) | 4,308.20±566.13 (2,555 to 6,337) | 20,067.90±4,231.25 (10,092 to 40,904) |

*Whole genome sequence and GWAS for feature selection.* The WGS of the human peripheral blood DNA were performed with an average read depth of 22× using a BGISEQ-500 sequencer (BGI Americas Corporation, Cambridge, MA, USA) of 150 bp paired-end reads [44]. The aligned and cleaned WGS data were mapped to the human reference genome (GRCh38/hg38) using Burrows-Wheeler Aligner software [47]. There were a total of 10,623,292 single nucleotide polymorphisms (SNPs) in the cohort with 935 subjects. For quality control, we removed genetic variants with missing rates larger than 5%, Hardy-Weinberg equilibrium exact test p-values less than $10^{-4}$, and minor allele frequency (MAF) less than 5%. Individuals with a missing rate larger than 20% were also excluded. Since subjects from two races were enrolled, principal component analysis (PCA) was applied to the genotypes and generated principal component scores (PCs) to perform population stratification or admixture [48]. In addition, the age, weight, height, and first 10 PCs were used as covariates in GWAS [17,41,42].

The genome-wide association analyses were performed to test the association between each of three phenotypes and SNPs from WGS. Suppose that there are $N$ subjects in the analyses. Let $y_i$ be the value of the $i$-th subject for a phenotype and $g_i$ be the genotype for the $i$-th subject, where $g_i$ is the number of minor alleles that the subject carries at a SNP. We assume that there is a total of $C$ covariates and the covariates for $i$-th subject are $\{v_1^{(i)}, v_2^{(i)}, \cdots, v_C^{(i)}\}$. For each SNP, the linear regression model is used to examine the effect of a SNP on a phenotype: $E(y_i|g_i) = \alpha_0 + \alpha_1 v_1^{(i)} + \cdots + \alpha_C v_C^{(i)} + \beta g_i$, where $\beta$ is the effect size of the SNP on the phenotype after adjusting for the covariates. The aim is to test the null hypothesis that the SNP is not associated with the phenotype, which is equivalent to test $H_0: \beta = 0$. The score test statistic under this model is: $T_{score} = UV^{-1}U$, where $U = \sum_{i=1}^{N} \tilde{y}_i \tilde{g}_i$ and $V = \frac{1}{N} \sum_{i=1}^{N} \tilde{y}_i^2 \sum_{i=1}^{N} \tilde{g}_i^2$, $\tilde{y}_i$ and $\tilde{g}_i$ are the adjusted phenotype and genotype for $i$-th subject for the covariates, indicating that $\tilde{y}_i$ is the residual of $y_i$ under the linear regression model $y_i = \alpha_0 + \alpha_1 v_1^{(i)} + \cdots + \alpha_C v_C^{(i)} + \varepsilon_i$ and $\tilde{g}_i$ is the residual of $g_i$ under the linear regression model $g_i = \alpha_0 + \alpha_1 v_1^{(i)} + \cdots + \alpha_C v_C^{(i)} + \tau_i$. Under $H_0$, $T_{score}$ follows a standard normal distribution [49].

*DXA and DXA-derived imaging features.* For each subject, aBMD (g/cm$^2$) at various skeletal sites (lumbar spine, hip, forearm, and total body) and body composition (fat/lean mass) were measured by Hologic Discovery-A DXA (Hologic Inc., USA) located in Tulane Center for Biomedical Informatics and Genomics. For quality control purposes: the DXA machine was calibrated daily, and long-term precision was monitored by phantoms with a coefficient of variation ≤0.7% for spine aBMD and a coefficient of variation ≤1.0% for hip aBMD [50]. Mechanical malfunction, radiation quality, absorption coefficient, and tissue-equivalent materials were also checked and calibrated before the aBMD examination on a daily basis. The radiologist was licensed in the State of Louisiana and registered through the American Registry of Radiologic Technologists. The detailed DXA-derived imaging features are shown in Table S1.

## 3.2. MVAE-PoE

We now present our contribution which consists of a multi-view variational autoencoder (MVAE) with the product of expert (PoE) for multi-view information integration. We extend the VAE from single-view input into multi-view input fashion. Notably, as the fact that the product of Gaussian distributions is also a Gaussian distribution, we apply the PoE to generate the common latent space for the variation inference with an analytical solution.

Suppose that under the multi-view setting, we have the data in $M$ views, $x_1, x_2, \cdots, x_M$. For the data in $m$-th view ($m = 1, \cdots, M$), a nonlinear function implemented by a neural network is employed as the encoder, denoted as $q_{\phi_m}\left(z_m \big| x_m^{(i)}\right)$, where $\phi_m$ represents the learnable parameters of the nonlinear function for $m$-th view. For each encoder, we estimate the mean vector and the variance-covariance matrix of multivariate Gaussian distribution for the approximate posterior distribution, denoted as $\mu_m^{(i)}$ and $\Sigma_m^{(i)}$ for $i$-th subject, and we assume $\mu_m^{(i)} \in \mathbb{R}^D$ is a vector and $\Sigma_m^{(i)} \in \mathbb{R}^{D \times D}$ is a diagonal matrix where $D$ is the dimension of the latent space. In our implementation, we employ multi-layer perceptron (MLP) as the encoder. To guarantee the positivity of the covariance, the output of the MLP is denoted as the $\log \Sigma_m^{(i)}$ first and then is converted to $\Sigma_m^{(i)}$ using the exponential function. Formally, the encoder is defined in Eq. 4.

$$\begin{aligned} q_{\phi_m}\left(z_m \big| x_m^{(i)}\right) &= \mathcal{N}\left(\mu_m^{(i)}, \Sigma_m^{(i)}\right) \\ &= \frac{1}{(2\pi)^{D/2} \sqrt{\left|\Sigma_m^{(i)}\right|}} \exp\left(-\frac{1}{2}\left(z_m - \mu_m^{(i)}\right)^T \left(\Sigma_m^{(i)}\right)^{-1} \left(z_m - \mu_m^{(i)}\right)\right) \\ \mu_m^{(i)} &= MLP_m^\mu\left(x_m^{(i)}\right) \\ \Sigma_m^{(i)} &= \exp\left(MLP_m^\Sigma\left(x_m^{(i)}\right)\right) \end{aligned} \tag{4}$$

where $z_m$ is the latent variable extracted by $m$-th view with the dimension of $D \times 1$. $MLP_m^\mu$ and $MLP_m^\Sigma$ are the neural networks for calculating mean and covariance, respectively. Let $T_m^{(i)} = \left(\Sigma_m^{(i)}\right)^{-1}$, then the multivariate Gaussian distribution for $m$-th view is rewritten as Eq. 5.

$$\begin{aligned} q_{\phi_m}\left(z_m \big| x_m^{(i)}\right) &= \frac{1}{(2\pi)^{D/2} \sqrt{\left|\Sigma_m^{(i)}\right|}} \exp\left(-\frac{1}{2} z_m^T T_m^{(i)} z_m + \left(\mu_m^{(i)}\right)^T T_m^{(i)} z_m + \Delta_m^{(i)}\right) \\ \text{where } \Delta_m^{(i)} &= -\frac{1}{2}\left(\mu_m^{(i)}\right)^T T_m^{(i)} \mu_m^{(i)} - \frac{D}{2} \log 2\pi + \frac{1}{2} \log\left|T_m^{(i)}\right| \end{aligned} \tag{5}$$

A PoE models the target posterior distribution of the common latent variable from multi-view as the product of the individual posterior distribution of the latent variable from single-view. According to Eq. 5, $\Delta_m^{(i)}$ is not related to the latent variable $z_m$. Therefore, for the following analysis, $\Delta_m^{(i)}$ is considered as a constant. In our MVAE-PoE, the PoE generates the common latent variable $z$ using Eq. 6.

$$q_\phi\left(z \big| x_1^{(i)} \cdots x_M^{(i)}\right) = \frac{1}{M} \prod_{m=1}^{M} q_{\phi_m}\left(z_m \big| x_m^{(i)}\right) \tag{6}$$

That is, the multivariate Gaussian distribution of the common latent variable is defined by the product of the multivariate Gaussian distribution of the latent variable extracted by $m$-th view. According to [51], the approximated posterior distribution of the common latent variable, $z$, is shown in Eq. 7.

$$q_\phi\left(z\middle|x_1^{(i)}\cdots x_M^{(i)}\right) = \mathcal{N}\left(\mu_z^{(i)}, \Sigma_z^{(i)}\right),$$
$$\mu_z^{(i)} = \left(\sum_{m=1}^{M}\left(\mu_m^{(i)}\right)^T T_m^{(i)}\right)\left(\sum_{m=1}^{M} T_m^{(i)}\right)^{-1} \quad (7)$$
$$\Sigma_z^{(i)} = \left(\sum_{m=1}^{M} T_m^{(i)}\right)^{-1}$$

where $\mu_z^{(i)}$ and $\Sigma_z^{(i)}$ are the mean vector and variance-covariance matrix of the approximated posterior distribution of common latent variable for $i$-th subject. To make the neural network differentiable, we adopt the reparameterization trick [40,52] to reparametrize the mean vector and the diagonal variance-covariance matrix of the multivariate Gaussian distribution, as shown in Eq. 8.

$$z^{(i)} = \mu_z^{(i)} + \left(\Sigma_z^{(i)}\right)^{1/2} \odot \epsilon_z \quad (8)$$

where $\epsilon_z \sim \mathcal{N}(0, I)$ and $\odot$ indicates the element-wise product. For each view, we employed MLP layers as the decoder to restore the features, which is denoted as $MLP_m^{dec}$ for $m$-th view. The graphical architecture of the proposed MVAE-PoE model is shown in Figure 1.

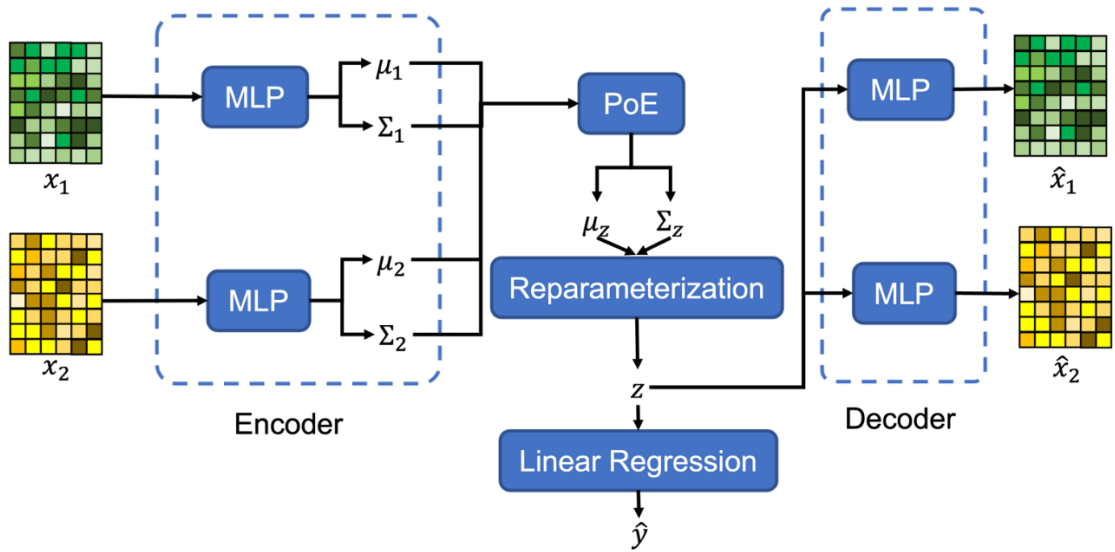

**Figure 1**. The graphical architecture of the proposed MVAE-PoE model.

*Loss function.* Using the multivariate Gaussian distribution, the ELOB for MVAE-PoE is derived in an explicit form, shown in Eq. 9.

$$\mathcal{L}(\theta, \phi; x_1, \cdots, x_M) = \sum_{i=1}^{N} \sum_{m=1}^{M} \mathbb{E}_{z \sim q_\phi(z|x_m^{(i)})} \log p_\theta \left( x_m^{(i)} | z \right) \\ - \sum_{i}^{N} D_{KL} \left( q_\phi \left( z | x_1^{(i)}, \cdots, x_M^{(i)} \right) \| p_\theta \left( z | x_1^{(i)}, \cdots, x_M^{(i)} \right) \right) \tag{9}$$

The first term in the RHS of Eq. 9 is defined as the cross-entropy between the reconstructed data and the original input, and the second term in the RHS of Eq.9 is the KL-divergence between the approximated posterior distribution and the true posterior distribution. The analytical form of the KL-divergence is the same as Eq. 3 since we employ the multivariate Gaussian distribution with an approximately diagonal variance-covariance as the ground truth. Thus, the close-form solution for the loss function is shown in Eq. 10.

$$\mathcal{L}(\theta, \phi; x_1, \cdots, x_M) = \sum_{i=1}^{N} \sum_{m=1}^{M} \left( x_m^{(i)} \log \left( \hat{x}_m^{(i)} \right) + \left( 1 - x_m^{(i)} \right) \log \left( 1 - \hat{x}_m^{(i)} \right) \right) \\ - \left( \frac{1}{2} \sum_{i=1}^{N} \sum_{d=1}^{D} \left( \left( \mu_d^{(i)} \right)^2 + \left( \sigma_d^{(i)} \right)^2 - \log \left( \sigma_d^{(i)} \right)^2 - 1 \right) \right) \tag{10}$$

where $\mu_d^{(i)}$ and $\left( \sigma_d^{(i)} \right)^2$ are the approximate mean and variance of the posterior distribution of $d$-th latent variable for $i$-th subject, and $\hat{x}_m^{(i)}$ represents the reconstructed feature vector for $i$-th subject from $m$-th view.

### 3.3. Model training and evaluation

20% of the subjects are randomly chosen as the test set, and the rest of the data are used as the training set. Predicting the proximal femoral strength is treated as a regression task. As shown in Figure 1, a linear regression model is employed to predict the proximal femoral strengths using the extracted latent variables, $z$.

For model evaluation, mean absolute error (MAE), mean absolute percentage error (MAPE), root mean squared error (RMSE) and $R^2$-score are employed. The definitions of MAE, MAPE, RMSE and $R^2$-score are shown in Eqs. 11-14.

$$MAE = \frac{1}{N} \sum_{i=1}^{N} |y_i - \hat{y}_i| \tag{11}$$

$$MAPE = \frac{1}{N} \sum_{i=1}^{N} \frac{|y_i - \hat{y}_i|}{y_i} \times 100\% \tag{12}$$

$$RMSE = \frac{1}{N} \sum_{i=1}^{N} \sqrt{(y_i - \hat{y}_i)^2} \tag{13}$$

$$R^2\text{-score} = 1 - \frac{\sum_{i=1}^{N}(y_i - \hat{y}_i)^2}{\sum_{i=1}^{N}\left(y_i - \sum_{i=1}^{N} y_i / N\right)^2} \tag{14}$$

where $y_i$ is the ground truth of the proximal femoral strength and $\hat{y}_i$ is the model prediction. A lower MAE/MAPE/RMSE and a higher $R^2$-score indicate better performance. According to Eqs. 11-13, 0 of MAE/MAPE/RMSE indicates the perfect match. According to Eq. 14, $R^2$-score ranges from $-\infty$ to 1, where 1 indicates the perfect match.

### 3.4. Interpretability of feature significance

Similar to [53], a leave-one-out technique is adopted to identify the feature significance in each view. A feature is significant if the performance of predicting proximal femoral strength decreases significantly when this feature is replaced by zero. By ranking the performance drops, the significance of the feature is obtained.

## 4. Experimental results

### 4.1. Data processing results

We performed GWAS analysis for testing the association between each of the three types of proximal femoral strengths, including LF, NLF and NLS, and each of the SNPs after quality control. Manhattan plots of these three types of proximal femoral strengths are depicted in Figure 2. Since the sample size of the LOS cohort was relatively small for genetic association studies, we expanded our search space to look at a much wider landscape of associations by selecting top 256 SNPs with the lowest p-values to extract the WGS features that are associated with each phenotype. These identified SNPs were used as WGS features for the downstream task. Meanwhile, 196 DXA-derived imaging features were employed.

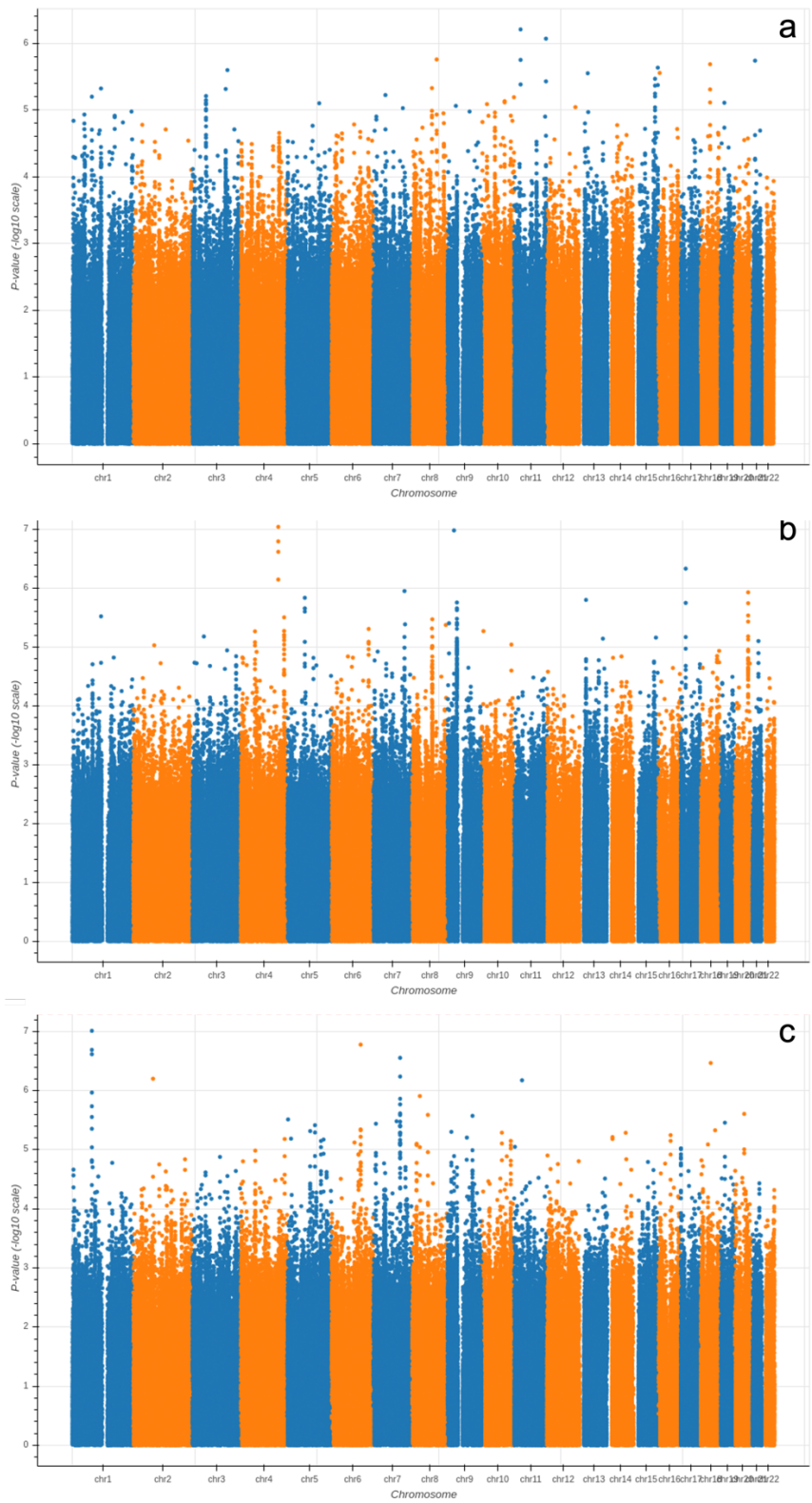

**Figure 2.** Manhattan plots of the GWAS results for a) LF; b) NLF; and c) NLS. The horizontal axis represents the chromosome index and the positions of the SNPs; while the vertical axis represents the p-value of GWAS results for each SNP.

## 4.2. Model performance for proximal femoral strength prediction

We trained and tested the MVAE-PoE model using our workstation with a NVIDIA RTX 3090 GPU and an Intel core I9 CPU. The designed models were implemented using TensorFlow 2.5. We performed the grid search to optimize hyperparameters, and the searching space included the number of MLP layers in both encoder and decoder: 1, 2, 3; the dimension of common latent space: 32, 48, 64, 128 or 256; and the number of hidden units for each MLP layer: 32, 48, 64, 128, or 256. Table 3 shows the best performance achieved using our proposed MVAE-PoE model for predicting three proximal femoral strengths. Also, we performed experiments using the different combinations of these three views to test the effectiveness of information fusion. We plotted the model prediction and the ground truth of the three FEA-computed proximal femoral strengths with the best performance in Figure 3.

**Table 3**. Fine-tuned best performance for the prediction of three proximal femoral strengths. The check marks in WGS features and DXA-derived features indicate that the corresponding view was used. The symbol ↑ indicates that higher is better and the symbol ↓ indicates that lower is better. If only one view was enrolled, then MVAE-PoE was degraded into a standard VAE model. For each type of the proximal femoral strength, the performance is sorted by $R^2$-score. The bold values indicate the achieved best performance.

| Phenotype | WGS | DXA | Number of MLP layers | Number of hidden units | Dimension of latent space | $R^2$-score ↑ | RMSE ↓ | MAE ↓ | MAPE ↓ |
|---|---|---|---|---|---|---|---|---|---|
| LF  | ✓ | ✓ | 3 | 48  | 128 | **0.5569** | **468.77** | **355.57** | **18.04%** |
|     |   | ✓ | 2 | 256 | 48  | 0.4866 | 504.56 | 388.65 | 19.66% |
|     | ✓ |   | 3 | 128 | 128 | 0.3317 | 575.68 | 453.11 | 24.47% |
| NLF | ✓ | ✓ | 2 | 128 | 128 | **0.5726** | **363.58** | **284.32** | **6.84%** |
|     |   | ✓ | 2 | 64  | 48  | 0.4778 | 401.92 | 306.88 | 7.39% |
|     | ✓ |   | 2 | 128 | 128 | 0.2979 | 466.02 | 368.17 | 8.89% |
| NLS | ✓ | ✓ | 3 | 256 | 48  | **0.7107** | **1903.58** | **1441.42** | **7.95%** |
|     |   | ✓ | 3 | 32  | 48  | 0.6822 | 1995.16 | 1539.79 | 8.41% |
|     | ✓ |   | 2 | 256 | 64  | 0.2194 | 3126.87 | 2430.34 | 13.94% |

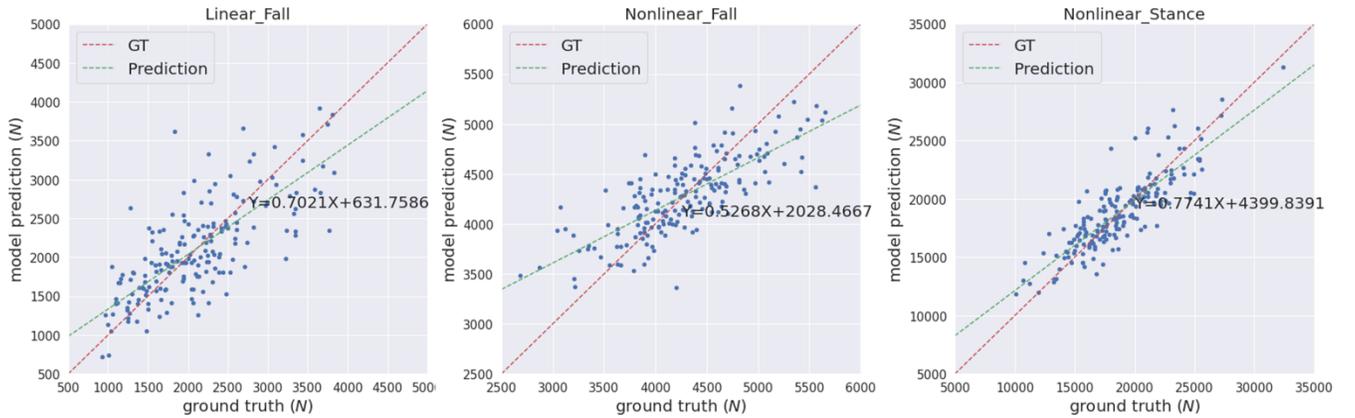

**Figure 3**. The predicted FEA-computed proximal femoral strengths and the ground truth (GT, in the horizontal axis) for LF, NLF and NLS. Each subject is represented by a blue dot. The vertical axis is the predicted strength. The red dashed line indicates a perfect match, and the green dashed line is the linear regression result of the prediction.

## 4.3. Performance comparison

We compared our MVAE-PoE model to other multi-view integration methods for prediction tasks. The tested models include:

- Multiview canonical correlation analysis (MCCA) [54]. MCCA extends the canonical correlation analysis (CCA) into multi-view settings. CCA is a typical subspace learning algorithm, aiming at finding the pairs of projections from different views with the maximum correlations. For more than 2 views, MCCA optimizes the sum of pairwise correlations.
- Kernel CCA (KCCA) [55]. KCCA is based on MCCA, however, it adds a centered Gram matrix to perform the nonlinear transformation on the input data.
- Kernel generalized CCA (KGCCA) [56]. KGCCA extends KCCA with a priori-defined graph connections between different views.
- Sparse CCA (SCCA) [57]: SCCA is a method for penalized CCA, which computes a rank-K approximation for a set of matrices and generates the sparse vectors for feature representation and interpretation.
- Multiview adversarial autoencoder (AAE) [58]. One limitation of the variational autoencoder is that the prior distribution and posterior distribution are required to be pre-defined, and the KL-divergence is required to be differentiable. The AAE can use arbitrary priors to train the autoencoder.

For the above algorithms, a linear regression estimator was applied to perform the prediction task using the extracted latent variables. The overall performance comparison for predicting the three proximal femoral strengths is shown in Table 4. For the compared algorithms, the grid search was also performed to find the best hyperparameters.

**Table 4**. Comparison of hip fracture load prediction between existing multi-view information extraction algorithms and the proposed MVAE-PoE. For each algorithm, the WGS features, and DXA-derived image features were used. Only the results with the best performance achieved by different algorithms are listed.

| Phenotype | Method | $R^2$-score ↑ | RMSE ↓ | MAE ↓ | MAPE ↓ |
|---|---|---|---|---|---|
| LF | MCCA | 0.2901 | 596.34 | 440.39 | 22.93% |
| | KCCA | 0.5212 | 489.77 | 391.58 | 20.43% |
| | KGCCA | 0.4215 | 538.35 | 423.28 | 22.34% |
| | SCCA | 0.5346 | 469.62 | 371.79 | 19.26% |
| | AAE | 0.4563 | 519.26 | 403.58 | 20.68% |
| | MVAE-PoE | **0.5569** | **468.77** | **355.57** | **18.04%** |
| NLF | MCCA | 0.2352 | 486.30 | 372.19 | 8.94% |
| | KCCA | 0.4095 | 427.31 | 337.09 | 8.13% |
| | KGCCA | 0.2541 | 480.23 | 365.32 | 8.76% |
| | SCCA | 0.4758 | 402.60 | 309.34 | 7.48% |
| | AAE | 0.5466 | 374.50 | 302.68 | 7.22% |
| | MVAE-PoE | **0.5726** | **363.58** | **284.32** | **6.84%** |
| NLS | MCCA | 0.3980 | 2739.70 | 1872.63 | 10.14% |
| | KCCA | 0.5958 | 2244.82 | 1700.97 | 9.38% |
| | KGCCA | 0.5178 | 2451.87 | 1796.35 | 9.86% |
| | SCCA | 0.6652 | 2043.08 | 1584.97 | 8.71% |
| | AAE | 0.5998 | 2238.89 | 1820.15 | 10.27% |
| | MVAE-PoE | **0.7107** | **1903.58** | **1441.42** | **7.95%** |

## 5. Discussion

### 5.1. Model performance

Integrating information from two views significantly improved the performance of predicting proximal femoral strength. According to Table 3, the proposed MVAE-PoE model achieved its best performance for LF, NLF and NLS prediction using WGS features and DXA-derived imaging features. For example, the proposed model improved the $R^2$-score to 0.5569 compared with 0.4866 using DXA features alone for LF prediction.

DXA-derived imaging features are significantly more important than the WGS features in terms of the prediction performance. For LF, NLF and NLS, using DXA features alone, the designed models achieved the MAPEs of 18.04%, 6.84%, and 7.95%, respectively. Integrating DXA features with WGS features, the MAPEs were lowered by 1.62%, 0.55% and 0.46%, respectively. This finding was consistent with clinical practice that DXA-derived imaging features correlate with bone weakness and fragility fracture [10], with site-specific DXA explaining approximately 55% of the variability in predicting proximal femoral strengths [59].

Proximal femoral strengths depend on WGS features; however, using WGS features alone, the model does not generate satisfactory prediction results. For predicting LF, using only WGS features increased the MAPE from 18.04% to 24.47%. For predicting NLF, using only WGS increased the RMSE from 363.58 to 466.02.

Compared to other multi-view information extraction models, the proposed MVAE-PoE achieved the best performance for predicting all types of proximal femoral strengths. The MCCA, KCCA, KGCCA and SCCA are four machine learning-based methods and the AAE is a deep learning-based method. For the MCCA, KCCA, KGCCA and SCCA, we trained these models with different dimensions of latent variables; for KCCA and KGCCA, we further tested the linear, polynomial and radial basis function (RBF) kernels. Even with tremendous hyperparameter fine-tuning, these machine learning-based methods didn't generate better performance than the designed MVAE-PoE models. For the AAE model, we employed the same grid search settings. However, the achieved MAPEs were 20.68%, 7.22% and 10.27% for LF, NLF and NLS prediction, which indicated inferior performance than MVAE-PoE. The CCA-based methods have been commonly used in data fusion or integration; however, CCA-based methods treat the modalities as linearly and multivariately correlated without considering the direction of the linear relationship [60]. In this study, we demonstrate that MVAE-PoE enables more useful and generalizable representations by capturing the abstract relationship between the views for downstream tasks such as prediction tasks.

**5.2. Feature importance analysis**

We applied the leave-one-out method to determine the feature importance. The leave-one-out indicated that we replaced one specific feature by zero for each subject in the test set when evaluating this feature, and the replaced features are named as zero-filled features. We compared the MAE changes between using the raw features and the zero-filled features. If the MAE between the GT and the model prediction increased significantly, then the evaluated feature was a significant feature. For each model, we listed the top 15 most important features in Figure 4.

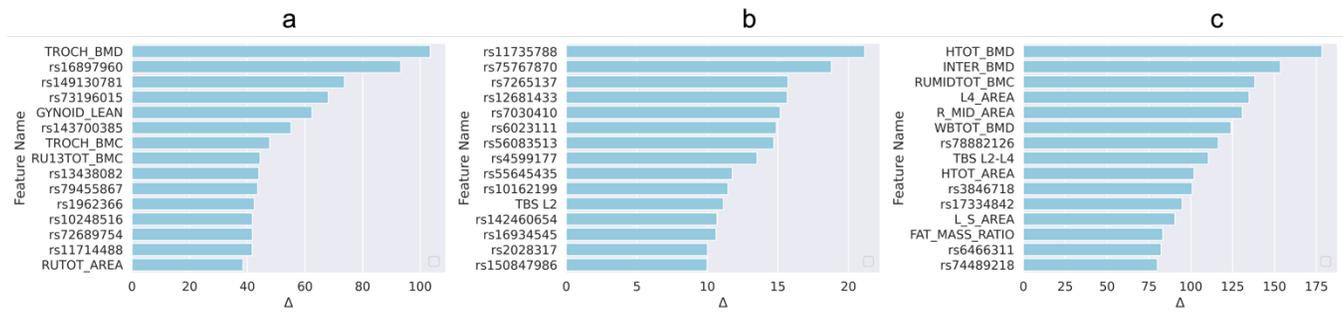

**Figure 4**. Ranked feature importance for a) predicting LF; b) predicting NLF; and c) predicting NLS. Feature significance was determined by MAE changes between using raw features and zero-filled features. The vertical axis indicates the feature names, where WGS features are annotated by rsid, and the DXA are annotated by the abbreviations. Detailed explanations for the DXA features are shown in Table S1.

For the prediction of LF, according to Figure 4 (a), 10 of the top 15 most important features were WGS features, and 5 feature was a the DXA feature. The trochanter BMD (TROCH_BMD) was the most important DXA feature. Trochanteric BMD is associated with trochanteric fracture in the elderly and is among the best predictors of femoral strength [61]. For NLF prediction, according to Figure 4 (b), one of the top 15 most significant features were DXA-derived features and the remaining 14 were WGS features. For NLS prediction, according to Figure 4 (c), 10 of the top 15 most significant features were DXA-derived features. This was consistent with the previous findings that DXA features explained approximately 55% of the proximal femoral strength while the proximal femoral strength was also influenced by genetics [20].

For each important WGS feature, we mapped the SNPs into the corresponding genes. The correspondingly associated clinical traits that were reported in the GWAS Catalog between each mapped gene and clinical traits are shown in Table 5. For LF, According to the meta-analysis using 339,224 subjects from 125 subjects including African Americans and Caucasians, TSPAN12 show a positive correlation with BMI [62]. COX6C and CAPG showed a strong correlation with body-shape indices on subjects from UK Biobank datasets [63]. For NLF, the detected most important genes, CAPG, also showed a strong correlation with body fat distribution [64]. For NLS, ERBB4 was associated with obesity on subjects from UK Biobank [65], which contained 339,244 individuals.

**Table 5.** Previously reported SNPs associated with bone-related clinical traits. The corresponding gene is listed.

| Phenotype | SNP | Nearest Gene | Clinical traits |
|---|---|---|---|
| LF | rs120932767 | TSPAN12 | BMI |
|  | rs16897960 | COX6C | Body-shape index |
| NLF | rs142460654 | CAPG | Body-shape index, Body Fat |
| NLS | rs17334842 | ERBB4 | Obesity |

### 5.3. Clinical application

Our proposed research not only addressed issues related to multi-view information fusion, but also leveraged the value of widely used DXA with information provided by genetic markers for predicting proximal femoral strength. Therefore, this study has the potential to significantly impact both research and clinical practice. In the AGES-Reykjavik data set, Fleps et al. demonstrated that using FEA-computed hip fracture load to predict hip fracture was better than using total femoral aBMD only [23]. In addition, genetic markers are important for identifying subjects at risk of hip fracture through effects on proximal femoral strength [20,22,66]. Due to their biological nature, genetic factors may also control nano-level bone mechanical properties and may further facilitate the prediction of bone strength and the assessment of hip fracture risk.

The most significant scientific impact of this study is the development and validation of the first comprehensive and accurate model for patient-specific assessment of predicting proximal femoral strength using multi-view information fusion by deep learning. Our multi-view deep learning-based model incorporates WGS features and DXA-derived imaging features, which are directly or indirectly related to proximal femoral strength and hip fracture. Deep learning-based techniques can automatically extract features and build accurate prediction models. Further, using the leave-one-out technique, the designed models are highly interpretable, leading to the identification of specific factors predictive of proximal femoral strengths.

The most practical clinical impact is the development and validation of an interpretable prediction model for proximal femoral strength using WGS features and DXA-derived image features, rather than using QCT. It is difficult to implement QCT-based femoral strength and hip fracture risk assessment in clinical practice due to the high radiation dosage and limited availability of QCT-based FEA. Lochmüller et al. suggested that clinical assessment of femoral fracture risk should preferably rely on femoral DXA [59]. Our results suggest that there is a strong potential for using a combination of DXA and genetic markers to develop practical models for hip fracture risk assessment in the future.

### 6. Conclusion

In this paper, we presented a multi-view variational autoencoder for latent feature extraction and the product of the multivariate Gaussian distributions for information fusion. Our model is based on the theory that the product of the multivariate Gaussian distributions is also a multivariate Gaussian distribution. Also, we assumed the latent distribution of the prior probability is represented by a diagonal variance-covariance matrix and thus induced the closed form of the KL-divergence and ELOB for variational inference and model training. We applied the designed MVAE-PoE to prediction of FEA-computed proximal femoral strength. For LF, NLF and NLS prediction, using WGS features, and DXA-derived images were important and beneficial. The proposed model achieved MAPEs of 18.04%, 6.84% and 7.95% for predicting LF, NLF and NLS, exceeding other existing methods significantly. From a technique perspective, our designed model can be applied to other multi-view information integration tasks for both regression and classification tasks. From a clinical perspective, though this tool is not a substitute for QCT-based FEA, it would make improved assessment of hip fracture risk more widely available while avoiding the increased radiation dosage from QCT.

**Acknowledgments**

This research was supported in part by grants from the National Institutes of Health, USA (P20GM109036, R01AR069055, U19AG055373, R01AG061917, R01AR27065, R01AG028832, R01AR46197, R01AR064140 and M01RR00585) and NASA Johnson Space Center, USA contracts NNJ12HC91P and NNJ15HP23P. It was also supported in part by a seed grant from Michigan Technological University Institute of Computing and Cybersystems, a graduate fellowship from Michigan Technological University Health Research Institute and a graduate fellowship from Portage Health Foundation.

# Supplementary Materials

**Table S1.** DXA-derived imaging features. LEAN: grams of lean tissue; FAT: grams of fat tissue; PFAT: percentage of fat tissue; BMD: bone mineral density in gram/cm$^2$; BMC: bone mineral content in gram; MASS, mass in grams of the corresponding region; AREA: area of the corresponding region in cm$^2$; VOLUME: volume of the corresponding region in cm$^3$.

| Feature name | Index | Description |
|---|---|---|
| ANDROID_GYNOID_RATIO | 1 | android-gynoid percent fat ratio |
| ANDROID_FAT/LEAN/MASS/PFAT | 2-5 | android bone |
| APPENDAGE_LEAN_MASS_HEIGHT_2 | 6 | appendage lean mass/height$^2$ |
| APPENDAGE_PURE_LEAN_HEIGHT_2 | 7 | appendage pure lean mass/height$^2$ |
| ARM_LENGTH | 8 | length of arm |
| BODY_MASS_INDEX | 9 | body mass index |
| BODY_WIDTH | 10 | body width in cm |
| CAVITY_WIDTH | 11 | cavity width in cm |
| FAT_MASS_HEIGHT_SQUARED | 12 | FMI, a measure of relative fat content |
| FAT_MASS_RATIO | 13 | Men: 64 – (20 x height/waist circumference) = RFM |
| GYNOID_FAT/LEAN/MASS/PFAT | 14-17 | gynoid bone |
| HEAD_AREA/BMC/BMD/FAT/LEAN/MASS/PFAT | 18-24 | femur head |
| HTOT_AREA/BMC/BMD | 25-27 | total hip |
| INTER_AREA/BMC/BMD | 28-30 | inter-trochanter bone |
| L1/L2/L3/L4/L5_AREA/BMC/BMD | 31-44 | L1/L2/L3/L4/L5 bone from spine |
| LARM_AREA/BMC/BMD/FAT/LEAN/MASS/PFAT | 45-51 | left arm region |
| LEAN_MASS_HEIGHT_SQUARED | 52 | lean mass/height2 |
| LLEG_AREA/BMC/BMD/FAT/LEAN/MASS/PFAT | 53-59 | bone of left leg |
| LRIB_AREA/BMC/BMD | 60-62 | bones of left rib |
| L_S_AREA/BMC/BMD | 63-65 | |
| NECK_AREA/BMC/BMD | 66-68 | bone of femur neck |
| OUTER_WALL_WIDTH | 69 | |
| PELV_AREA/BMC/BMD | 70-72 | |
| PURE_LEAN_HEIGHT_SQUARED | 73 | |
| RARM_AREA/BMC/BMD/FAT/LEAN/MASS/PFAT | 74-80 | bone of right arm |
| RLEG_AREA/BMC/BMD | 81-83 | bone of right leg |
| ROI_HEIGHT/TYPE/WIDTH | 84-86 | forearm region of interest |
| RRIB_AREA/BMC/BMD | 87-89 | bones of right rib |
| RTOT_AREA/BMC/BMD | 90-92 | bone of total radius |
| RU13TOT_AREA/BMC/BMD | 93-95 | bone of 1/3 distal radius and ulna |
| RUMIDTOT_AREA/BMC/BMD | 96-98 | bone of mid distal radius and ulna |
| RUTOT_AREA/BMC/BMD | 99-101 | bone of radius and ulna |
| RUUDTOT_AREA/BMC/BMD | 102-104 | bone of ultra distal radius and ulna |
| R_13_AREA/BMC/BMD | 105-107 | bone of 1/3 distal radius |
| R_LEG_FAT/LEAN/MASS/PFAT | 108-111 | bone of right leg |
| R_MID_AREA/BMC/BMD | 112-114 | bone of mid distal radius |
| R_UD_AREA/BMC/BMD | 115-117 | bone of ultra distal radius bone |
| SAT_AREA/MASS/VOLUME | 118-120 | subcutaneous adipose tissue |
| SUBCU_FAT_CORRECTION | 121 | correction for subcutaneous fat |
| SUBTOT_AREA/BMC/BMD/FAT/LEAN/MASS/PFAT | 122-128 | total bone area |
| TAT_AREA/MASS/VOLUME | 129-131 | total adipose tissue |
| TBS L1/L2/L3/L4/L1-L4/L2-L4 | 132-137 | trabecular bone score |
| TOTAL_FAT/FAT_MASS/LEAN/LEAN_MASS/MASS/PFAT | 138-144 | bones of total android |
| TOT_AREA/BMC/BMD | 145-147 | bones of total spine |
| TORCH_AREA/BMC/BMD | 148-150 | bone of trochanter |
| TRUNK_LIMB_FAT_MASS_RATIO | 151 | trunk/limb fat mass ratio of fat |
| TRUNK_FAT/LEAN/MASS/PFAT | 152-155 | bones of the trunk |
| T_S_AREA/BMC/BMD | 156-158 | bone of thoracic region. |
| UTOT_AREA/BMC/BMD | 159-161 | bone of total ulna |
| U_13_AREA/BMC/BMD | 162-164 | bone of 1/3 distal ulna |
| U_MID_AREA/BMC/BMD | 165-167 | bone of mid distal ulna |
| U_UD_AREA/BMC/BMD | 168-170 | bone of ultra distal ulna |
| VFAT_BODY_FAT/LEAN/MASS/PFAT | 171-174 | visceral |
| VFAT_CAVITY_FAT/LEAN/MASS/PFAT | 175-178 | visceral adipose tissue of abdominal cavity |
| VFAT_OUTERWALL_FAT/LEAN/MASS/PFAT | 179-182 | visceral adipose tissue of abdominal wall |
| VFAT_VOLUME/MASS/AREA | 183-185 | visceral adipose tissue of abdominal volume |
| WAIST_CIRCUMFERENCE | 186 | circumference of waist |
| WARDS_AREA/BMC/BMD | 187-189 | ward's triangle bone |
| WBTOT_AREA/BMC/BMD/FAT/LEAN/MASS/PFAT | 190-196 | whole body |